\pdfoutput=1 

\documentclass[final,3p,times]{elsarticle}

\usepackage{amssymb}
\usepackage{amsmath,amsfonts}
\usepackage{array}
\usepackage{subfigure}
\usepackage{textcomp}
\usepackage{stfloats}
\usepackage{url}
\usepackage{verbatim}
\usepackage{graphicx}
\usepackage[noend]{algpseudocode}
\usepackage[linesnumbered,ruled]{algorithm2e} 
\usepackage{float}
\usepackage{booktabs,makecell,multirow}
\usepackage{booktabs}
\usepackage{threeparttable}
\biboptions{sort&compress}

\journal{}

\begin{document}

\begin{frontmatter}



\title{An Energy Optimized Specializing DAG Federated Learning based on Event Triggered Communication}


\author[HIT-CS]{Xiaofeng Xue}

\author[HIT-CS]{Haokun Mao}

\author[HIT-CS]{Qiong Li\corref{cor1}}
\ead{qiongli@hit.edu.cn}
\cortext[cor1]{Corresponding author}
\author[HIT-FLS]{Furong Huang}

\affiliation[HIT-CS]{organization={School of Cyberspace Science,Faculty of Computing, Harbin Institute of Technology},
            city={Harbin},
            country={China}
            }

\affiliation[HIT-FLS]{organization={School of International Studies, Harbin Institute of Technology},
	city={Harbin},
	country={China}
    }
    
\begin{abstract}
Specializing Directed Acyclic Graph Federated Learning(SDAGFL) is a new federated learning framework which updates model from the devices with similar data distribution through Directed Acyclic Graph Distributed Ledger Technology (DAG-DLT). SDAGFL has the advantage of personalization, resisting single point of failure and poisoning attack in fully decentralized federated learning. Because of these advantages, the SDAGFL is suitable for the federated learning in IoT scenario where the device is usually battery-powered. To promote the application of SDAGFL in IoT, we propose an energy optimized SDAGFL based event-triggered communication mechanism, called ESDAGFL. In ESDAGFL, the new model is broadcasted only when it is significantly changed. We evaluate the ESDAGFL on a clustered synthetically FEMNIST dataset and a dataset from texts by Shakespeare and Goethe's works. The experiment results show that our approach can reduce energy consumption by 33\% compared with SDAGFL, and realize the same balance between training accuracy and specialization as SDAGFL.
\end{abstract}




\end{frontmatter}


\section{Introduction}
\label{sec: Introduction}
With the Internet of Things(IoT) development, data have become more diversified and distributed, stimulating the demand for privacy and security \cite{Albrecht2016, Yi2021}. As a result, traditional centralized cloud-based machine learning has been challenged. A few machine learning techniques have been proposed to meet these challenges. A technique called Federated Learning (FL)\cite{Konecny2016} provides a promising solution that allows the clients to work together to build a global machine learning model without sharing the local data on their own devices. A typical federated learning framework consists of a server and some local clients. The clients in FL can access the same global model and train it using local data. Then, they update the trained local model to the server. Once the locally trained models are received, the server will aggregate them as a new global model and feed it back to the clients for the next local update. FL  will repeat this process several rounds until the global model converges or the number of repetitions meets the predetermined target.

In this way, the client can protect data privacy as FL  implements model training without collecting user data. Nevertheless, there are still some challenges in Federated Learning as follows:

\begin{itemize}
    \item \textbf{Device Heterogeneity:} In FL, the server must wait for all the selected clients to complete the local training before aggregating the local models. However, the computing power and network bandwidth vary from client to client\cite{Niknam2020, Chen2021}. The clients with higher computing power and network bandwidth can complete the local training faster than the others. Thus, the server will wait a long time until the slowest client to complete the local training before starting the next round. It leads to the performance bottleneck of FL\cite{Xu2022}.
    \item \textbf{Data Heterogeneity:} A significant difference between FL and other machine learning is that model's training in FL is completed on the clients with the data of Non-IID. The Non-IID data distribution will cause a significant decrease in the global model's accuracy. The exiting reach shows that the accuracy can be reduced by 55\% of the neural network using the highly skewed Non-IID dataset\cite{Zhao2018}. 
    \item \textbf{Single Point of Failure:} In the traditional FL, there is a single central server to aggregate the local model and publish the new global model. Therefore, the hacker can attack the central server to cause a single point of failure, leading to a performance decrease or even a failure of FL training.
    \item \textbf{Poisoning Attack:} The clients in FL are not all honest, and some clients may use the poisoned local data to train the model and then send the poisoned model to the server. However, the server cannot detect the poisoned model and will aggregate it into the global model. The poisoning attack will cause a decrease in the global model's accuracy\cite{Cao2019}.
\end{itemize}

To address the challenges mentioned above, distributed ledger technology (DLT) based on directed acyclic graph (DAG) with decentralized, high throughput and tamper-resistant advantages is introduced into the federated learning\cite{Schmid2020,Yuan2021,Cao2021,Beilharz2021}. Among these works, Specializing DAG Federated Learning (SDAGFL\footnote{We express Specializing DAG Federated Learning as SDAGFL in this work for  simplicity}) \cite{Beilharz2021} is the one that attracts more attention. In the SDAGFL, the participating clients use a DAG for the communication of models and an accuracy-biased random walk to obtain the models from other devices with similar data distribution to update their local model. It not only achieves overcoming the challenges of device heterogeneity, failure of a single point, and poisoning attack but also creates a balance between reaching a consensus on a generalized model and personalizing the model to the clients different from the traditional FL framework where all the participating clients training and reaching consensus for a global model together.
 
With the above advantages of the SDAGFL, it is suitable for the FL in a IoT scenario. However, some devices in the IoT scenario is usually powered by a battery with limited battery energy. Therefore, reducing energy consumption while maintaining training performance is a problem that needs to be solved to promote the application of SDAGFL in IoT.

In this work, we first analyze the energy consumption in SDAGFL. And we propose an Event-triggered communication based SDAGFL approach, called event-triggered SDAGFL (ESDAGFL), to reduce the energy consumption for SDAGFL. The main works of this article are summarized as follows:

\begin{enumerate} 
    \item We firstly analyze the energy consumption of the SDAGFL and give the energy consumption formula of SDAGFL. Then we give the energy optimization objective of SDAGFL.
    \item To promote the application of SDAGFL in IoT, we propose an optimization algorithm based on the event-triggered communication mechanism to optimize the energy consumption of SDAGFL. 
    \item We perform the simulation experiments on two datasets to verify our approach. The experimental results prove the convergence of our approach. Furthermore, the experiment results show that our approach can reach an equal balance between performance and specialization in SDAGFL. Our approach can reduce the energy consumption of SDAGFL.
\end{enumerate}

The rest of the paper is organized as follows. In Section \ref{sec: Related Work}, we introduce the basic concept of DAG distributed ledger, review the art of the federated learning framework based on DAG-DLT and show the advantages of SDAGFL over other frameworks. Then, we introduce the event-trigger mechanism. In Section \ref{sec: SDAGFL and ECA}, we analyze and formulate the energy consumption and give the optimization objective. What follows is that we propose an event-triggered communication-based SDAGFL approach in Section \ref{sec: Framework}. Numerical simulation results are presented in Section \ref{sec: Experiments}. Finally, we summarize the work and give possible future research in Section \ref{sec: Conclusion}.

\section{Related Work}
\label{sec: Related Work}
In this section, we will detail the basic concept of DAG-DLT and the four works of federated learning based on the DAG-DLT. Next, we analyze the advantages of SDAGFL compared with other works. Finally, we will introduce the event-triggered mechanism.
\subsection{DAG-DTL and Federated Learning based on DAG-DLT}
DAG-based Federated Learning is inspired by the DAG Distributed Ledger Technology (DLT). The DAG-DLT can achieve consensus Similar to the Nakamoto DLT \cite{Nakamoto2008}, which is also called blockchain, in a distributed system. The DAG-DLT uses a DAG data structure to store the transactions, allowing multiple transactions to be added to the ledger simultaneously. As a result, the DAG-DLT is considered suitable for the scenario that includes many distributed devices. The Tangle\cite{Popov2018} created by the IOTA foundation is a typical DAG-DLT designed for the devices of the Internet of Things to participate in a low-energy network. The Tangle has a higher Transaction Per Second (TPS) than the blockchain because the transactions in the Tangle can be confirmed within minutes. 

Because the Tangle has the advantages of high TPS, low energy usage, and decentralization, it has received the researchers' extensive attention. 

Four Tangle-based federated learning frameworks have been proposed as far as we know. Cao et al.\cite{Cao2021} deployed a federated learning framework with Tangle called DAG-FL. DAG-FL includes three layers: the federated learning layer, the DAG layer, and the application layer, and it implements federated learning with asynchronous training and anomaly detection features. Shuo Yuan et al. \cite{Yuan2021} proposed ChainsFL, a two-layer hierarchical federated learning framework combining the Tangle and blockchain. The ChainsFL consists of a sub-ledger, the Raft \cite{Ongaro2014}-based Hyperledger Fabric \cite{Androulaki2018}, deployed on edge nodes, and a Tangle-based master ledger. The ChainsFL has overcome the disadvantages of massive resource consumption and limited throughput in traditional blockchain-based FL. In addition, Schmid Robert et al. discussed the basic applicability of the combination of federated learning and the Tangle in \cite{Schmid2020}. They assumed a long-standing open network for continuous learning and development. In this framework, the tip selection algorithm based on Monte Carlo Markov Chain in the Tangle is used to implement model selection. The averaged model is used to update the model. The evaluation of this framework shows the high training accuracy and model-agnostic resistance against random poisoning and label-flipping attacks. Beilharz et al. further optimized the framework in \cite{Schmid2020}. They proposed an implicit model specialization framework for federated learning called Specializing DAG Federated Learning(SDAGFL). In SDAGFL, the implicit clustering of clients with a similar local dataset is achieved through the accuracy-biased random walk on DAG-DLT and Fedavg algorithm. 

The above four frameworks show that the integration of the Tangle with federated learning can implement asynchronous training and poisoning attack resistance. What is better than the other three works, SDAGFL further sufficiently uses the DAG data structure's feature, and an accuracy-biased random walk for model update realizes implicit clustering of the client. Thus, we think the SDAGFL, which uses adaptive device and data heterogeneity, resistant poisoning attack, and personal federated learning, is worthy of more study.

In particular, if not specified, the DAG-DLT in this work refers to the Tangle, and SDAGFL refers to the framework proposed in \cite{Beilharz2021}

\subsection{Event-triggered Communication Mechanism}
Event-triggered Communication Mechanism is proposed for the network control systems to reduce the frequency of communication \cite{Dimarogonas2012, Lemmon2010}. In an event-triggered mechanism-based system, the device will perform the predetermined operation only when it detects an event that meets the trigger conditions. The clients in the specializing DAG FL proposed by Beilharz et al. \cite{Beilharz2021} will search for a ``Reference Model'' after training to decide whether to broadcast the new model, which can be regarded as an event-triggered communication mechanism. However, it is not friendly for the devices in the IoT scenario where the battery is limited. 

The event-triggered communication mechanism has been introduced into the field of parallel machine learning recently, and some event-triggered Stochastic Gradient Descent (SGD) has been proposed. In the event-triggered SGD, the client computes the model's gradient and broadcasts the gradients that meet the trigger threshold to the neighbors. \cite{Ghosh2022,Nguyen2021,George2020,George2019,Kajiyama2018}. 

Inspired by event-triggered SGD, we utilize an event-triggered communication based on the parameter change to reduce the energy consumption of SDAGFL in our work.
\section{Specializing DAG Federated Learning and Energy Consumption Analysis}
\label{sec: SDAGFL and ECA}
\subsection{Specializing DAG Federated Learning}
In a IoT scenario, there are some devices participating in the federated learning  powered by the battery with limited power-energy. It is necessary to reduce the energy consumption of federated learning. The benefit of reducing the energy consumption is two folds. On the one hand, it can reduce the number of charging times and prolong the service life of the device. On the other hand, it is friendly to the environment. To solve this problem, we first introduce the SDAGFL in detail and then analyze the CPU energy consumption in the SDAGFL.

\begin{figure}[ht]
    \centering
    \includegraphics[width=85mm]{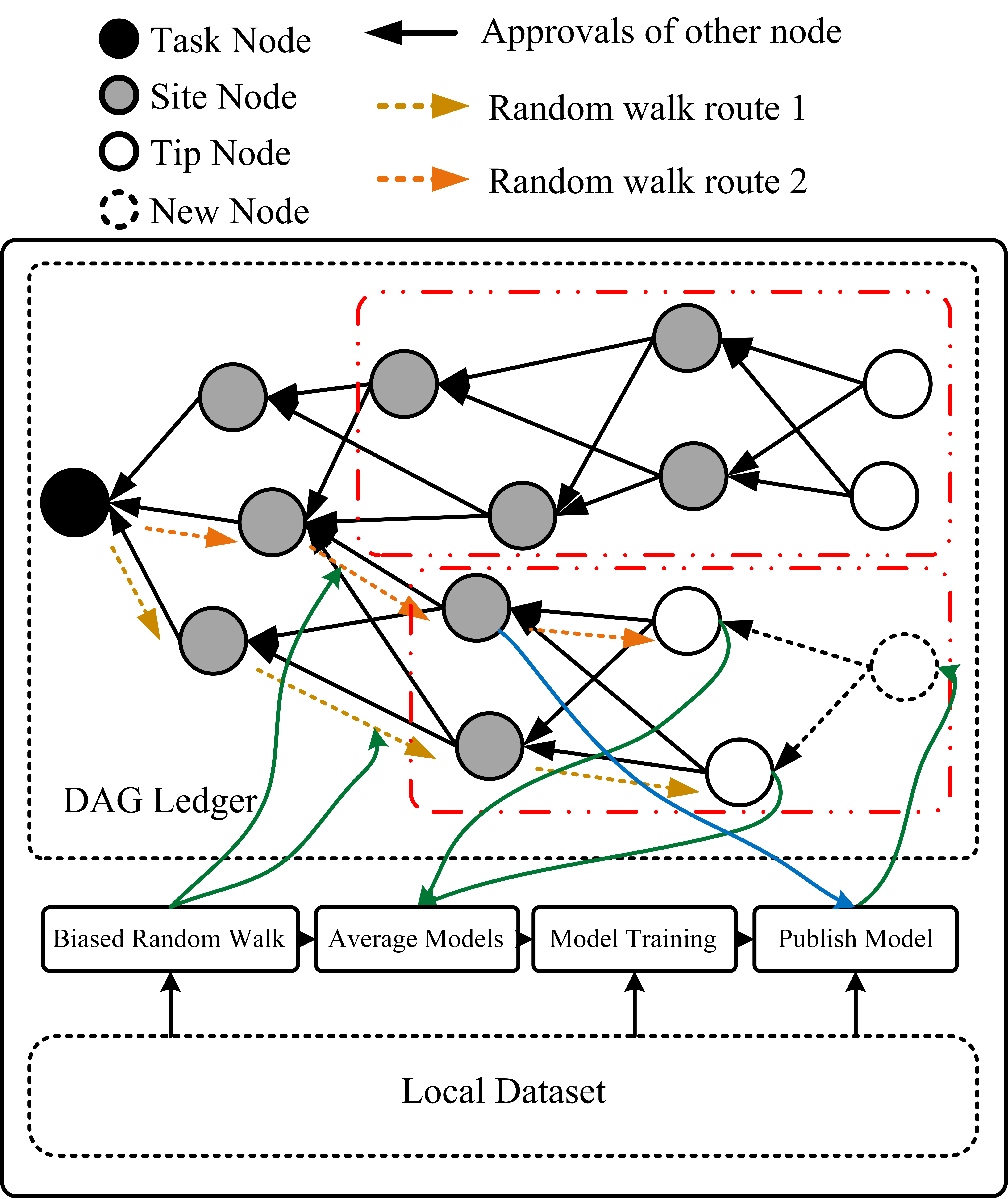}
    \caption{Schematic diagram of the basic principle of Specializing DAG Federated Learning}
    \label{Basic principle of Specializing DAG Federated Learning}
\end{figure}

The basic principle of SDAGFL is shown in Figure \ref{Basic principle of Specializing DAG Federated Learning}. There is a public DAG ledger shared by all the clients in SDAGFL. The nodes in SDAGFL are divided into four categories, and each node contains a complete model parameter.

\begin{enumerate}[a)]
    \item \textbf{Task Node:} The task node is a genesis node. It contains the basic training task and initial model parameter;
    \item \textbf{Site Node:} The site node is the basic node type. It contains the model published by the clients;
    \item \textbf{Tip Node:} The tip node is a special site node, which is not approved by other node;
    \item \textbf{New Node:} The new node contains the model that will be published.
\end{enumerate}

During the process of the SDAGFL, the clients will repeat the following five steps to update the model parameter:

\begin{enumerate}[Step 1]
    \item Each client performs two accuracy-biased random walks in algorithm \ref{algorithm:randomwalk} based on the
    local dataset until reaching to the tip node. We show the random walk route using the yellow dashed line in figure \ref{Basic principle of Specializing DAG Federated Learning}; \label{step:randomwalk}
    \item Then, the client averages the models in two tips obtained from step \ref{step:randomwalk}.\label{step:averagemodel}
    \item Next, the client trains the average above model using its local dataset, and then obtains a new model.
    \item In this step, the client will perform several accuracy-biased random walks to obtain the ``Reference Model''(the site node approved by the blue line in figure \ref{Basic principle of Specializing DAG Federated Learning}). 
    \item Finally, the client will compare the new model with the ``Reference Model'' to determine whether to publish the new model. If the loss, test using the local data, of new model is lower than the ``Reference Model'', the client will publish the new model to the public DAG ledger.
\end{enumerate}

\begin{algorithm}[H]
    \caption{Random Walk of Specializing DAG FL}
    \label{algorithm:randomwalk}
    $children$ $\gets$ GetChildren(Site Node $n$)\;
    $n=len(children)$\;
    initial $accuracy[n]$\;
    \For{$child[i]$ in children}
    {
        $accuracy[i] = EvaluateOnLocalData(child[i])$\;
    }
    \For{$accuracy[i]$ in $accuracy$}
    {
       $normalized[i]= \frac{accuracy[i] - max(accuracy)}{max(accuracy)-min(accuracy)}$\;
       $weight[i] = e ^{normalized}\times \alpha$ \;
    }
    \tcp{WeightChoice is a  weight-based child node selection function}
    nextnode = WeightChoice(weight) \;
    RandomWalk(nextnode)\;
\end{algorithm}

With the increase of the DAG ledger, the site node trained by the clients with independent identical distributed data will be clustered in the same group. As we can see, the site nodes in Figure \ref{Basic principle of Specializing DAG Federated Learning} are clustered into two groups (red dashed box).

\subsection{Energy Consumption Analysis}
After understanding the basic principle of SDAGFL, we will analyze its energy consumption. The energy consumption in SDAGFL can be divided into four parts: Client obtain two tip nodes, Model Aggregation, Model Training, client obtain and compare with the reference model.

\textbf{Client obtains two tip nodes:} The energy consumption of this process is decided by the depth of the ledger and the number of children node approval the site node. We define the depth of the ledger as $d$, and the number of CPU cycles required to test the model accuracy is $f_i^w$. The $f_i$ is defined as the CPU frequency of the client $i$. In addition, we assume that the average number of site node's children is $l$ and the CPU chipset's effective capacitance coefficient is $C_i$. The energy consumption can be expressed as follows according \cite{Tran2019a}

\begin{equation}
    E_i^{tip} = C_i \times 2 \times {d l f_i^w} \times {f_i^2}
\end{equation}

\textbf{Client aggregates the model in tip node:} In this process, we define the number of CPU cycles to aggregate the model as $f_i^a$. In thus, the energy consumption in model aggregation is
\begin{equation}
    E_i^{aggregation} = C_i \times {f_i^a} \times {f_i^2}
\end{equation}

\textbf{Client trains the model:} The energy cost of model training depends on the size of the dataset. The number of CPU cycles for training one sample of dataset is defined as $f_i^t$, and the training batch size and number of batch is defined as $b_i$ and $b_{num}$. The training epoch is defined as $e$. The energy consumption express as 
\begin{equation}
    E_i^{training} = C_i \times b_i b_{num} e f_i^t \times {f_i^2} 
\end{equation} 

\textbf{Client obtains and compares with the reference model:} To obtain the  reference model, the client need to compute the confidence and rating of each site node. The confidence is the number of the site node selected during the multiple random walks. And the rating is the number of nodes that directly and indirectly approved by every site node. Thus, the energy cost of this process contains two aspects: confidence computing and rating computing. Thus, the energy consumption of the client obtaining the ``Reference Model'' can be expressed as follows.
\begin{equation}  
    E_i^{reference} = C_i \times r \times d (lf_i^w + f_i^c + lf_i^r) \times f_i^2
\end{equation}

Where $r$ is the number of random walk and it is $5$ in SDAGFL. $f_i^r$ is the number of CPU cycles for computing each node's rating. $f_i^c$ is the is the number of CPU cycles for computing each node's confidence.

Through the above analysis, we can get the energy consumption by the client $i$ for one time model update as follows:

\begin{equation}
    \begin{aligned}
        E_i^{total} &= E_i^{tip} + E_i^{aggregation} + E_i^{training} +E_i^{reference} \\
        &= C_i \times \{2 d l f_i^w + f_i^a + b_i b_{num} e f_i^t + r d  (lf_i^w + f_i^c + lf_i^r) \} \times f_i^2
    \end{aligned}
\label{total consumption}
\end{equation}

We note that the process in which the client obtains two tip nodes is an essential process for SDAGFL. This process is related to the depth $d$ of the ledger and it can be optimized by taking a random walk from a depth of 15 to 20 transactions from tip nodes\cite{Beilharz2021}. In addition, the model's size determines the model aggregation process's energy consumption. Therefore, $f_i^a$ in the process of model aggregation is a fixed parameter in a specific federated learning task. As for the process of model training, energy consumption is connected to the training batch. To be fair and to simplify the problem, we let each device execute the same batch. To compute the rating and the confidence of the site node, the client must traverse the ledger from the Task node several times. Although reducing the walk depth $d$ is benefited to reduce the energy consumption, this process needs to be repeatedly executed $r$ times which also result the huge energy consumption. 

All things considered, we can reduce the energy consumption of SDAGFL by optimizing the process of the client obtain and compare with the reference model. Thus, the optimization objective of this work is formulated as follows:
\begin{equation}
    Minimize [\mathbb E, \underset {\omega \in {{\mathbb{R}}^{d}}}{\mathop{\min }}\,f(\omega)]   
\end{equation}

Where $\mathbb E = \sum{_{j=1}^T}{\sum{_{i=1}^N} C_i \times \{r d  (lf_i^w + f_i^c + lf_i^r) \} \times f_i^2 } $, and the $\underset {\omega \in {{\mathbb{R}}^{d}}}{\mathop{\min }}\,f(\omega)$ stands for minimizing the federated learning training loss function.

In other words, the optimized objective can be expressed as minimizing the energy consumption of the client while minimizing the federated learning training loss function.

\section{Framework}
\label{sec: Framework}
\begin{figure*}[htbp]
    \centering
    \includegraphics[width=150mm]{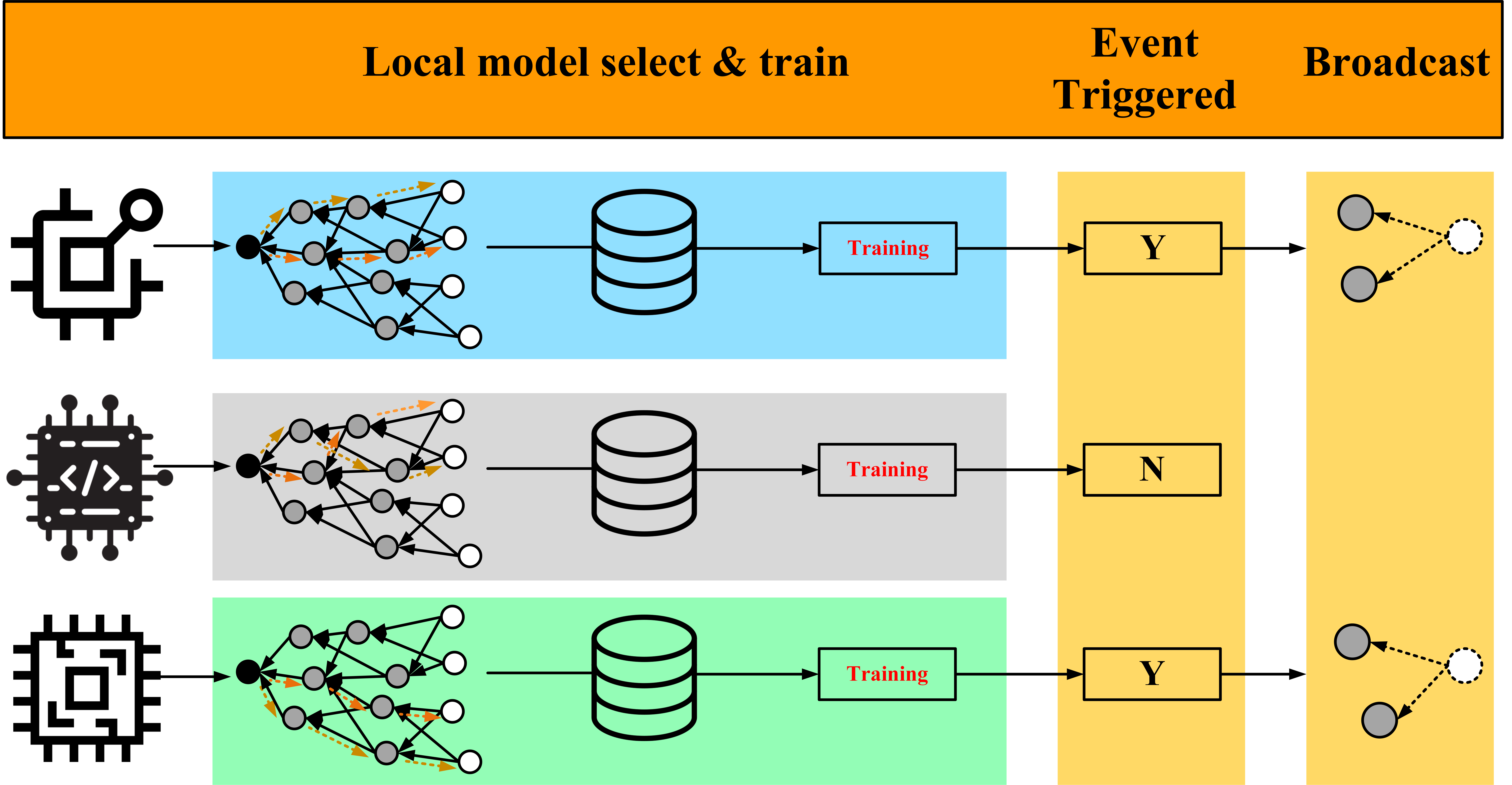}
    \caption{Framework of Event-triggered Specializing DAG FL}
    \label{Overview}
\end{figure*}

The aim of this work is to train multiple similar but identical models with minimizing the energy consumption and federated learning training loss function from the analysis in section \ref{sec: SDAGFL and ECA}. Therefore, we propose an Event-triggered communication-based SDAGFL(ESDAGFL) to meet this goal. The workflow of SDAGFL is shown in Figure \ref{Overview}.

We first define that the clients participating in the federated learning have the features as follows:
\begin{itemize}
    \item \textbf{Data are private to the clients:}In ESDAGFL, the dataset is stored in the spatially distributed clients, and it is private to all clients. And there is no data exchange among the clients. 
    \item \textbf{Data are Non-IID for all clients:} Data is non-independent and identically distributed (Non-IID) among the all clients participating in the ESDAGFL;
    \item \textbf{Data are IID for clients in the same cluster:} The site node published by the clients with a similar data  distribution feature would be clustered into a ``implicit community''. So, we can regard that the data distribution among the clients clustered in the same community is independently identical distribution.
    \item \textbf{Honest and Malicious client's behavior:} The honest clients always comply the rules of the SDAGFL and use their complete local datasets to participate in the training. Once the trained model  meets the broadcast condition, the client will broadcast it in time. On the contrary, the malicious clients do not comply any rules.
    \item \textbf{Synchronicity:} Although, the ledger is not synchronized in the vast majority of reality cases, we think that the performance of the ESDAGFL would not be affected by the ledger is synchronous or not. The reason is that whether the client publish the new model is only depend on the average model and trained new model. In thus, we define that the ledger on each client is synchronized in real-time to analyze the performance of our approach for conveniently.
\end{itemize} 

Like general federated aggregation algorithm such as FedAvg, we use the same supervised objective function suggested. The optimized model is defined as $\omega$. The predictive loss function on data points $\left(x_i,y_i\right)$ can be expressed as:${{f}_{i}}\left( w \right)=\left( {{x }_{i}},{{y}_{i}},w \right)$. The clients with the same data distribution will be clustered in multiple implicit communities. For every cluster, we assume that the number of clients in each community is $M$, and each client with $k$ data points $|k|=n_i$. Therefore, each model optimization is defined as:
\begin{equation}
    \underset {\omega \in {{\mathbb{R}}^{d}}}{\mathop{\min }}\,f(\omega)
\end{equation}
where
\begin{equation}
    f(\omega )=\sum\limits_{i=0}^{M}{\frac{{{n}_{m}}}{n}\times {{F}_{i}}(\omega )} 
\end{equation}
and
\begin{equation}
    {{F}_{i}}(\omega )=\frac{1}{{{n}_{i}}}\sum\limits_{j}^{{{n}_{i}}}{{{f}_{j}}(\omega )}
\end{equation}

In our approach, every model optimization is defined as using a distributed gradient descent to minimize the loss function on the local dataset. The client first obtains the aggregated model $\omega_{avg}^i = Avg({{\omega }_{tip1}},{{\omega }_{tip2}})$ with two tip nodes through random walk. Then, it computes the local gradient $\nabla {{F}_{n}}(\omega_{avg})$ and performs $\tau$ local model parameter updates using the local dataset to train the new model ${{\omega }_{new}}^i$. This process can be expressed as equation (\ref{sgd}).

\begin{equation}
    \label{sgd}
    {{\omega }_{new}^i}=\omega_{avg}^i -\eta \nabla {{F}_{n}}(\omega_{avg})
\end{equation}

Next, we can define the model parameter change rate between the trained new model and the input averaged model by equation (\ref{ETC-Delta})
\begin{equation}
    \label{ETC-Delta}
    \Delta^i = {||\Delta\omega^i||_2}/{||\omega_{avg}^i||_2}
\end{equation}

Where,
\begin{equation}
    \Delta\omega^i = | \omega_{new}^i -\omega_{avg}^i |
\end{equation}

We define that the trained new model $\omega_{new}^i$ will be broadcasted once the model parameter change rate is equal to or greater than the trigger threshold. 

\begin{equation}
    \label{trigger-time}
    \Delta^i \geq trigger_{threshold}
\end{equation}

In detail, the Event-triggered SDAGFL is summarized in Algorithm \ref{Alg-ETC}.

\begin{algorithm} 
    \caption{ESDAGFL Algorithm}
    \label{Alg-ETC} 
    \KwIn{learning rate $lr$ , total number of local updates $\tau$}
    \KwOut{New Model:$\omega_{new}$} 
    ($\omega_1$,$\omega_2$)$\gets$Random Walk of SDAGFL fo client i\;
    $\omega_{avg}$ = ${(\omega_1+\omega_2)}/{2}$\;
    \For{$h=0,1,2,...,\tau-1$}
    {
        $\omega_{new}^i = \omega_{avg} - \eta g_i(x^i_h, \xi^i) $\;
    }
    $\Delta\omega^i = \omega_{new}^i - \omega_{avg}$\; 
    Compute $\Delta^i = {||\Delta\omega||_2}/{||\omega_{avg}||_2}$ \; 
    \If{$\Delta^i \geq threshold_{tirgger}$}
    {
        return $\omega_{new}^i$ \;
    }   
\end{algorithm}
\section{Experiment Results}
\label{sec: Experiments}
In this section, we will introduce the evaluation result of ESDAGFL based on the framework\footnote{\url{https://github.com/osmhpi/federated-learning-dag}} proposed by Beilharz et al.\cite{Beilharz2021} with Pytorch. The parameters of the evaluation platform used in the experiment are shown in table \ref{tab:platform}.

\begin{table}[H]
    \centering
    \label{tab:platform}
    \caption{Parameters of the Experiment Platform}
    \begin{threeparttable}
        \begin{tabular}{cccccc}
            \toprule
                &Parameter \\
            \midrule
            OS & Ubuntu 20.04 \\
            CPU & Intel$^\circledR $Core$^\text{TM}$ i9-10980XE @3.0GHz$\times$36 \\
            Memory & 32GB \\
            \bottomrule
        \end{tabular}
    \end{threeparttable}  
\end{table}

\subsection{Experimental Setting}
This section introduces the experimental setting of ESDAGFL. The datasets and models used in the experiment are described in \ref{sec:dataset}, and 
the fixed training hyperparameters setting is shown in \ref{sec:hyperparameters}.
\subsubsection{Datasets and Models}
\label{sec:dataset}
We evaluated our approach on two training tasks. We evaluated a handwriting recognition task on the FMNIST-clustered dataset and a next character prediction task on the Poets dataset. The training and test datasets have a ratio of 9:1 for each client. 
\begin{itemize}
    \item \textbf{Handwriting Recognition Task:}
    \begin{itemize}
        \item \textbf{Dataset:} FMNIST-clustered dataset is a synthetically clustered version of Federated extended MNIST (FMNIST) built by LEAF\cite{Caldas2019}. In this dataset, the 28x28 pixel handwriting dataset is divided into three disjoint classes :[0,1,2,3], [4,5,6], [7,8,9], and the number of clients is the same in every class.
        \item \textbf{Model:} For the handwriting recognition task, a Convolutional Neural Net (CNN) is used. It contains two convolution layers and two fully connected layers. The kernel size of the convolution layer is 5 with the RELU activation function. Each convolution layer is followed by a max pooling layer with pool size and stride length 2. The two fully connected layers contain 2048 and 10 neurons, respectively.
    \end{itemize}
    \item \textbf{Next Character Prediction Task:}
    \begin{itemize}
        \item \textbf{Dataset:} The Poets dataset is used to evaluate the performance of our approach for the next character prediction task. The Poets dataset consists of an English dataset from William Shakespeare's works and a German dataset from Goethe's plays. The English and German datasets have an equal number of samples and are put into different clusters.
        \item \textbf{Model:} For the next character prediction task, the model first maps each character to an embedding of dimension 8, calculated from the 80 character sequence. Then the model passes each character through a Long Short-Term Memory(LSTM) consisting of two layers with 256 units each. Finally, there is a dense layer for prediction.
    \end{itemize}
\end{itemize}

\subsubsection{Training Hyperparameters Setting}
\label{sec:hyperparameters}
The training hyperparameters setting is shown in Table \ref{hyperparameters}. 

\begin{table}[H]
    \centering
    \caption{Training Hyperparameters Setting of the Experiment}
    \begin{tabular}{cccccc}
        \toprule
        Parameters &FMNIST-clustered&Poets \\
        \midrule
        Local epochs $E$ & 1 & 1 \\
        Local batches $B$ & 10 & 200 \\
        Batch size $B_{size}$ & 10 & 10 \\
        Learning rate $lr$ & 0.05 & 0.8 \\
        Clusters per round & 10 & 10 \\
        \bottomrule
    \end{tabular}
    \label{hyperparameters}
\end{table}

\subsection{Experimental Results}
Different from the traditional FL, ESDAGFL is a fully asynchronous federated learning framework without any central server. It realizes the training target through each client runs the training process continuously and independently as long as its resources permit. Therefore, the concept of the rounds is introduced to enable a better demonstration of the experimental results.

Moreover, the clients are assigned to the same process and in our simulation experiments. Thus, the energy consumption can be considered directly proportional to the training time. We use the energy consumption of the process ``client obtains two tip nodes'' as an example. The energy consumption can be expressed as equation (\ref{energy consumption}).
\begin{equation}
    \label{energy consumption}
    E_i^{tip} = C_i \times 2 \times {d l f_i^w} \times {f_i^2} 
\end{equation}

And the time cost can be expressed as equation (\ref{time cost}).
\begin{equation}
    \label{time cost}
    T_i^{tip} = 2 \times {d l f_i^w}/{f_i}
\end{equation}

From the equation (\ref{energy consumption}) and (\ref{time cost}), we can find that both energy consumption and time cost are proportional to $f_i^w$. Thus, we evaluated the time consumption instead of the energy consumption to show the advantages of our method.

In the experiment, the $threshold_{trigger} $ is set to $0.008$ for FMNIST-clustered dataset and $0.12$ for Poets datasets. The evaluation of training accuracy and loss was done every five rounds using the test dataset with 5\% of all clients randomly selected. We regard the work in \cite{Beilharz2021} as the baseline, and we evaluated our ESDAGFL approach on three metrics: training accuracy and loss, implicit specialization, and training time cost.

\subsubsection{Training accuracy and loss Evaluation} 
\label{train-acc-loss evaluation}
Training accuracy and loss are two standard metrics to evaluate the model performance. The accuracy is the percentage of correct predictions, and the loss is cross-entropy loss.

\begin{figure}[h]
    \centering  
    \subfigure[Training Accuracy]{
        \label{ETC-fmnist-acc}
        \includegraphics[width=80mm]{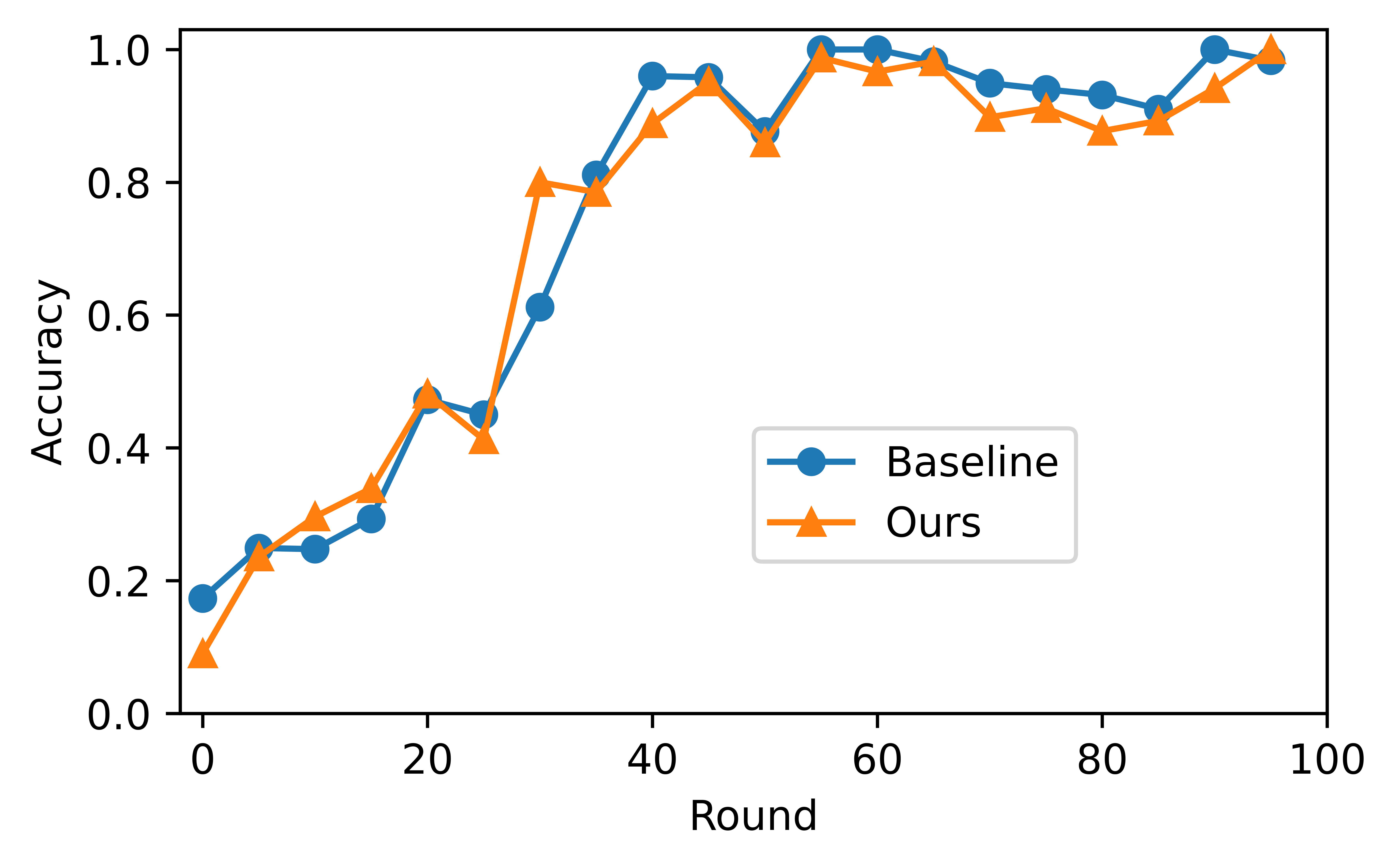}}
    \subfigure[Cross Entropy Loss]{
        \label{ETC-fmnist-loss}
        \includegraphics[width=80mm]{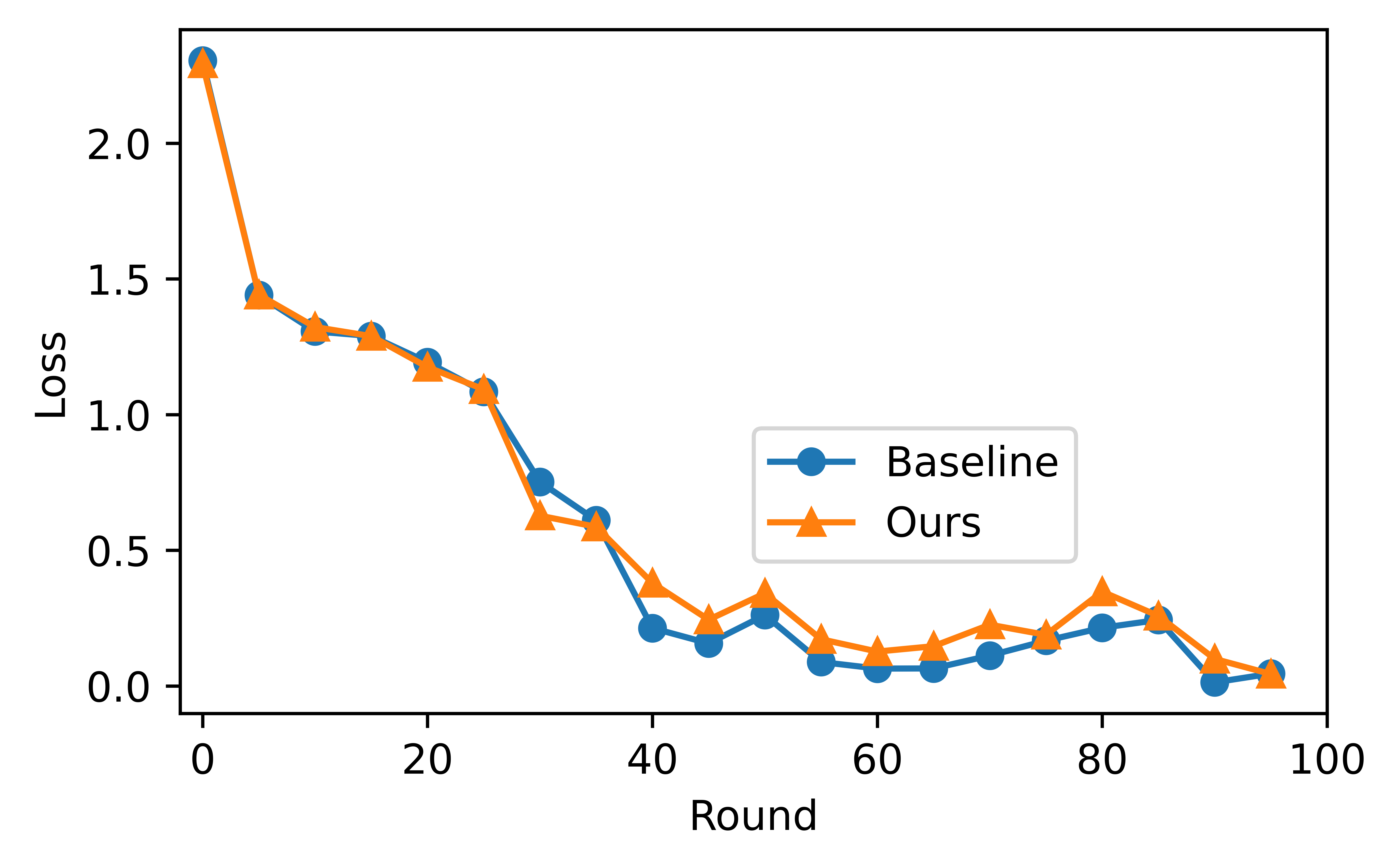}
        }
    \caption{Our approach's average training accuracy and loss compared with SDAGFL in the FMNIST-clustered dataset.}
    \label{ETC-fmnist-acc-loss}
\end{figure}

Firstly, we evaluated the average accuracy and loss on the test dataset for the handwriting recognition task. The evaluation results are shown as Figure \ref{ETC-fmnist-acc-loss}.

The evaluation result shows that the ESDAGFL approach achieves similar accuracy and approximate loss as the baseline. The experimental result demonstrates the applicability of our approach to the written recognition task.

\begin{figure}[H]
    \centering  
    \subfigure[Training Accuracy]{
        \label{ETC-Poets-acc}
        \includegraphics[width=80mm]{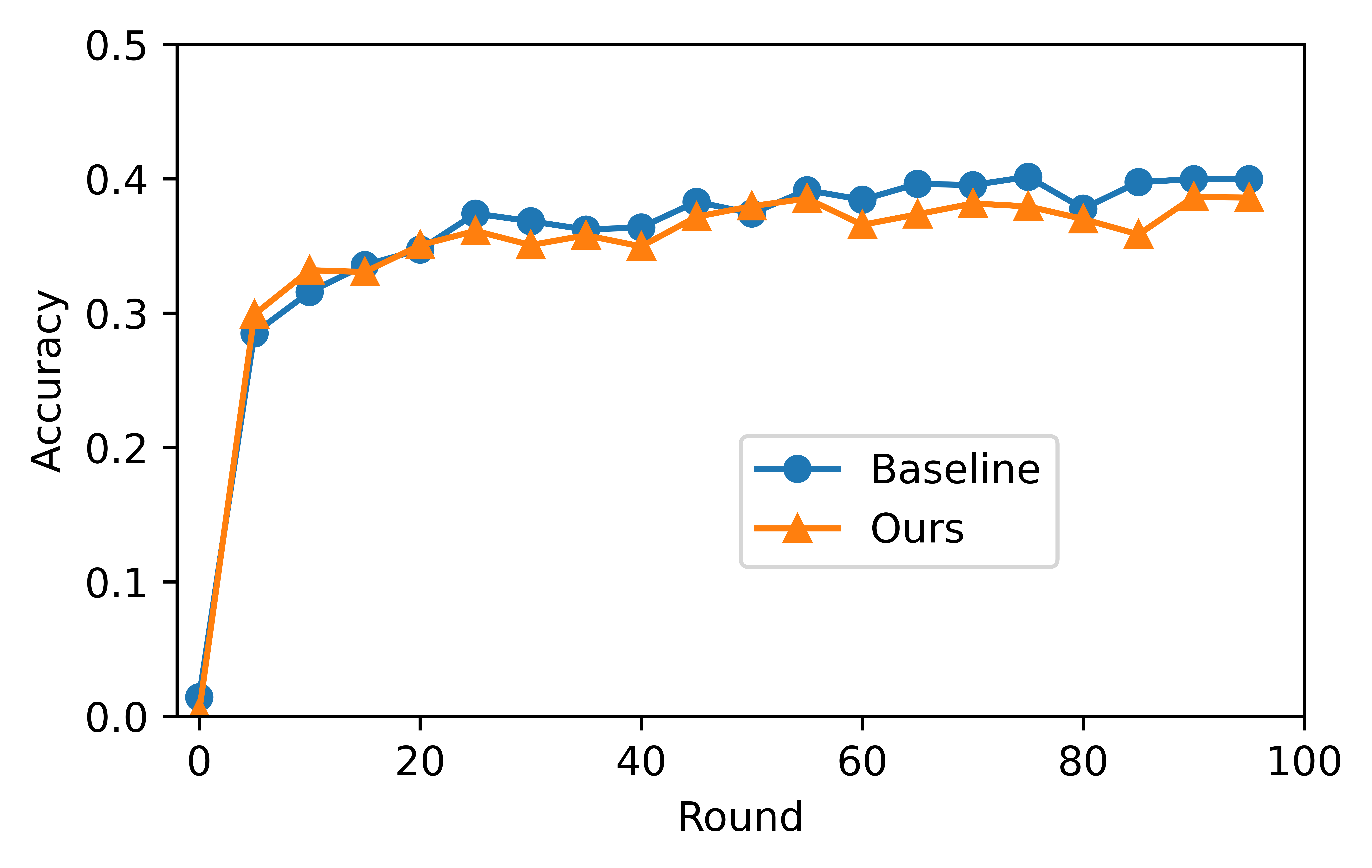}}
    \subfigure[Cross Entropy Loss]{
        \label{ETC-Poets-loss}
        \includegraphics[width=80mm]{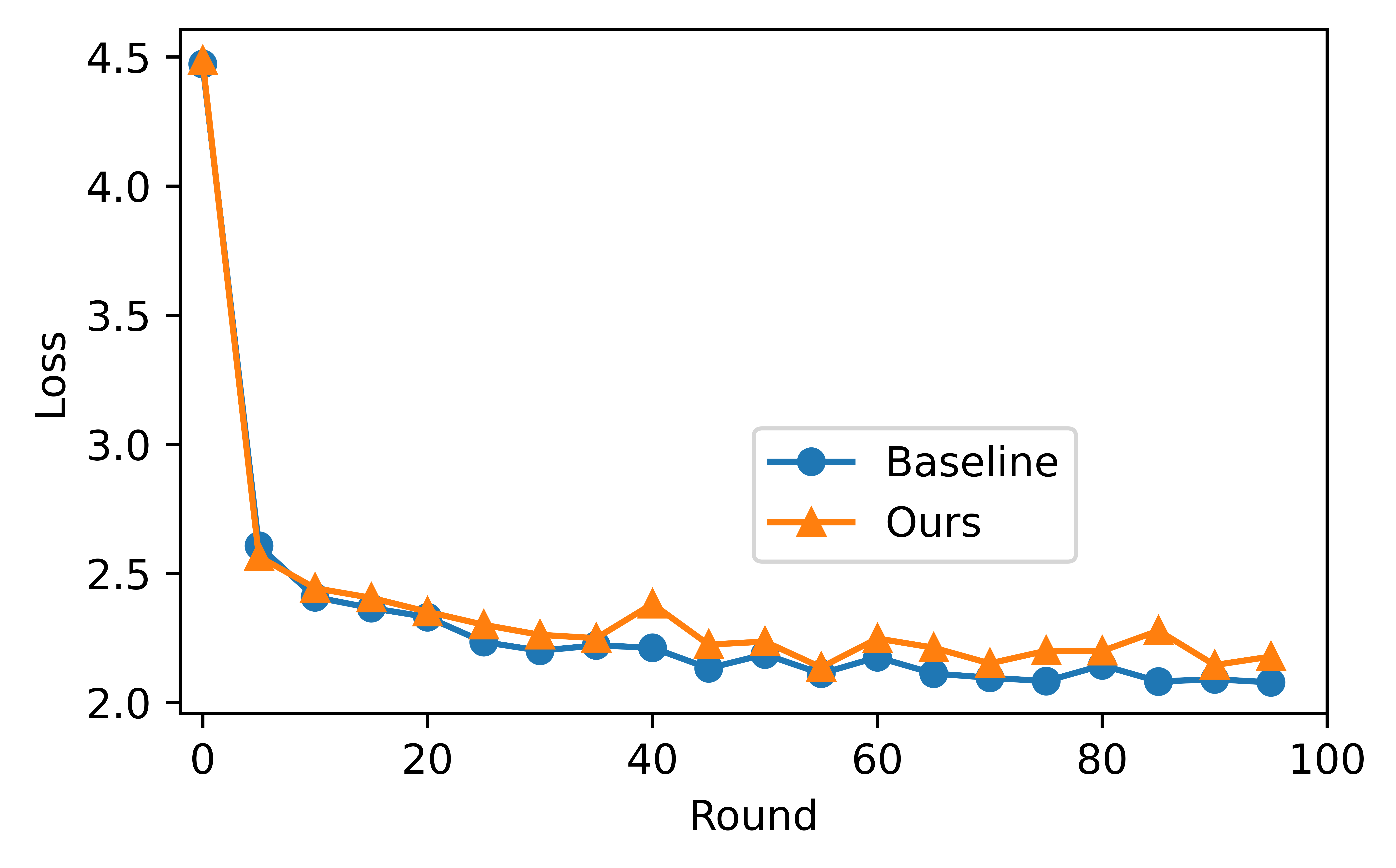}
        }
    \caption{Our approach's average training accuracy and loss compared with SDAGFL evaluated on the Poets dataset.}
    \label{ETC-Poets-acc-loss}
\end{figure}

Then, we evaluated the next character prediction task on the Poets dataset to further illustrate the applicability of our approach for different models. Figure \ref{ETC-Poets-acc-loss} shows our approach's average accuracy and loss in the Poets dataset. The experiment results show that our approach equally applies to the Poets dataset. 

\textbf{Convergence of ESDAGFL} As the new models are only published if they are changed when the honest clients' training for the model is positive, the federated training will converge to an expected direction. Figure \ref{ETC-fmnist-acc-loss} and \ref{ETC-Poets-acc-loss} show the convergence rate of ESDAGFL on the FMNIST-clustered and Poets datasets. Multiple experiment results show that the training process converges to nearly the same result. It confirmed that our approach could converge to a reasonable training accuracy. In addition, Whether the client publishes the trained new model is decided by the norm change of the model. Therefore, our method reduces the possibility of federated learning training falling into local optimization.

\subsubsection{Implicit specialization Evaluation}
\label{implicit-specialization evaluation} 
In this part, we evaluated the implicit specialization of the ESDAGFL. There are three metrics to evaluate the implicit specialization of the ESDAGFL. The first is the modularity $m\in[-\frac{1}{2},1]$, which is a metric of the community segmentation in the community discovery algorithm. The more closely connected the nodes within a community and the more sparsely the connections between the communities are, and the greater the modularity. The second metric is the number of modules which indicates the partitioning of all clients. The number of modules should be an appropriate value. The final metric is approval pureness, which measures the probability that a client approves the nodes published by other clients in the same cluster. 

Firstly, we evaluated the modularity and number of modules on the FMNIST-clustered dataset. The only FMNIST-clustered dataset that is chosen is because that the FMNIST-clustered dataset has more classes and is easier to visualize.

\begin{figure}[ht]
    \centering  
    \subfigure[Modularity]{
        \label{ETC-fmnist-Modularity}
        \includegraphics[width=80mm]{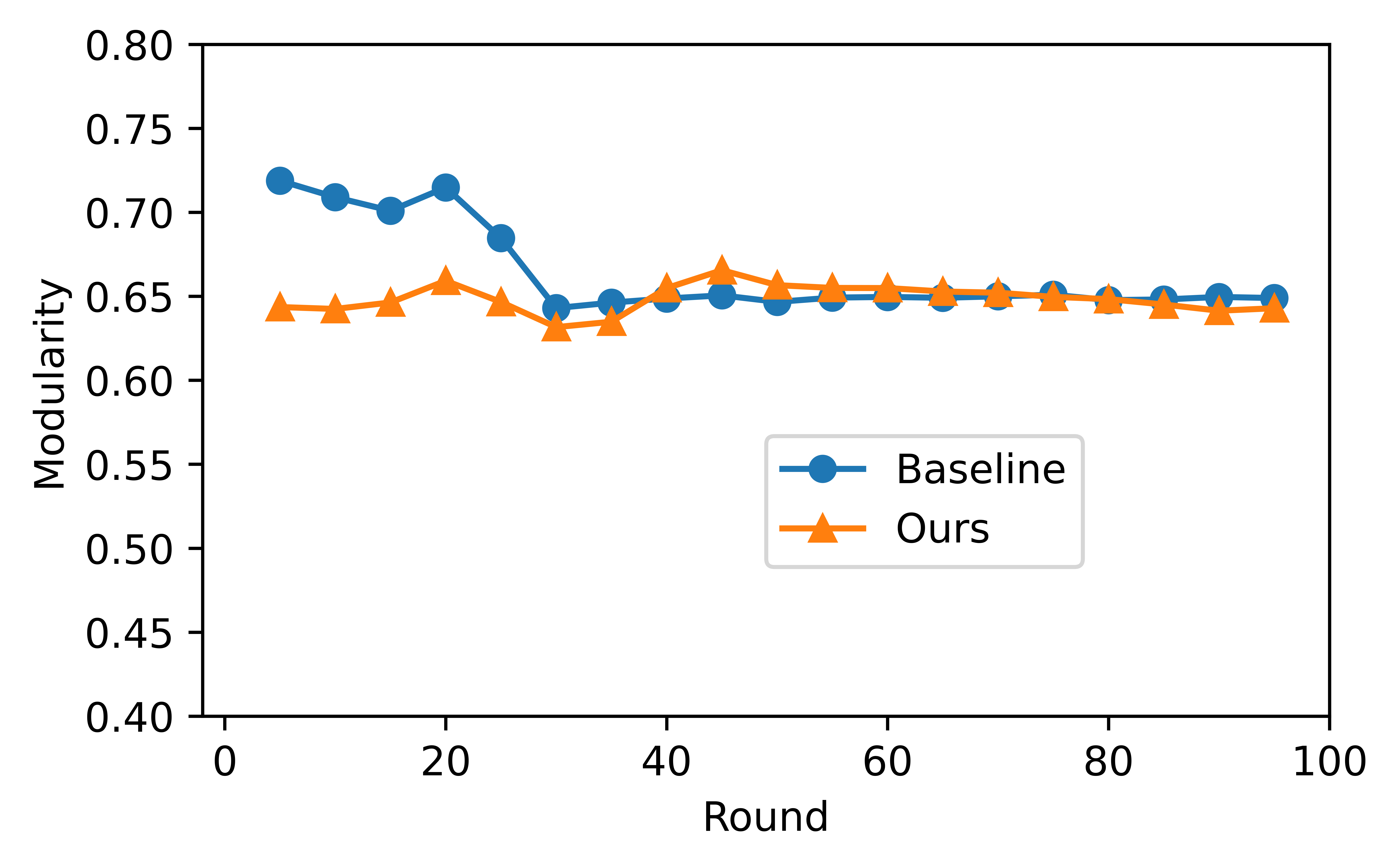}}
    \subfigure[Modules]{
        \label{ETC-fmnist-Modules}
        \includegraphics[width=80mm]{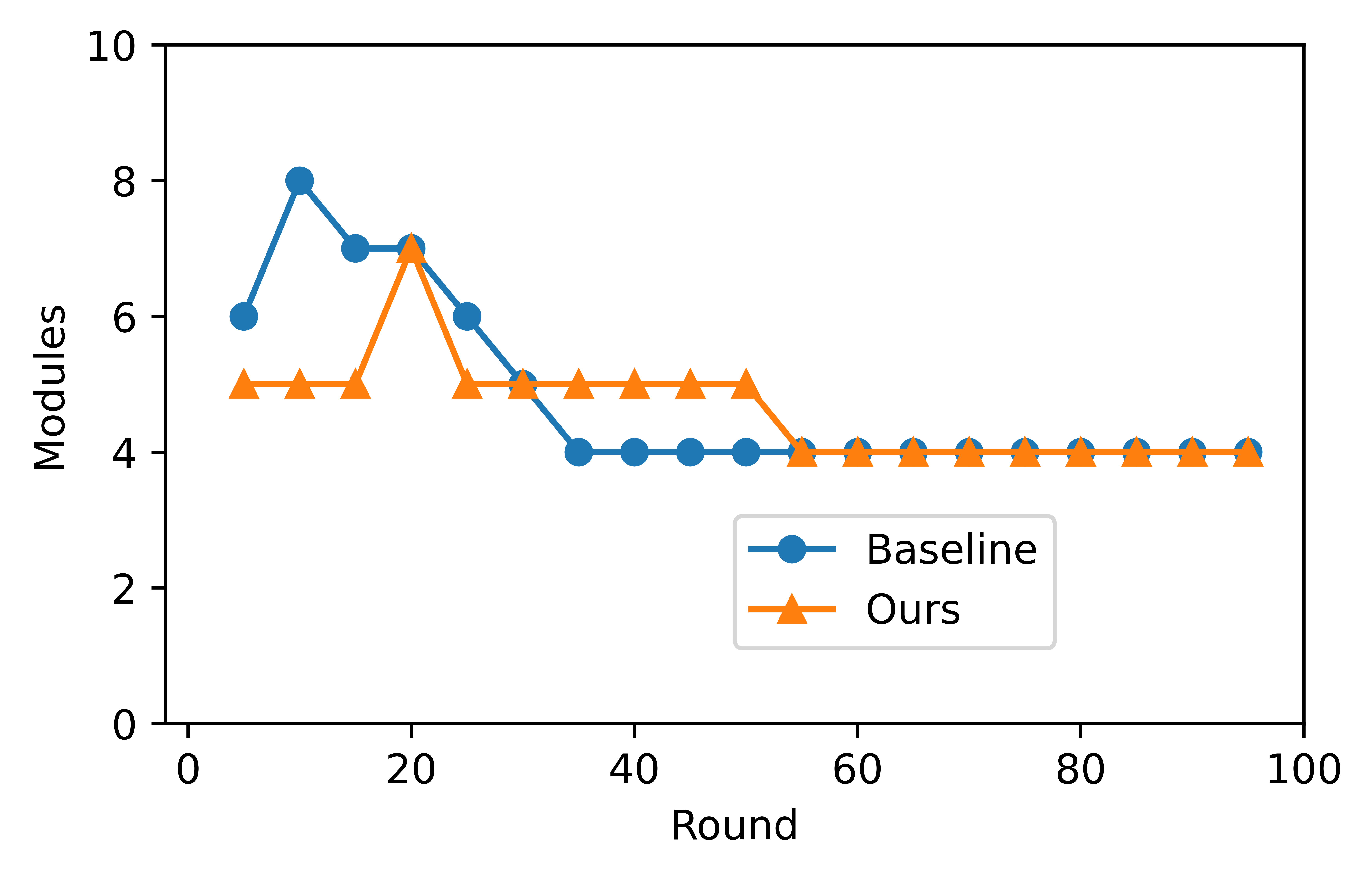}
        }
    \caption{The Modularity and Modules variation in our approach and baseline evaluated on the FMNIST-clustered dataset}
    \label{ETC-fmnist-Modularity-Modules}
\end{figure}

Figure \ref{ETC-fmnist-Modularity-Modules} shows the performance of our approach in the implicit specialization. Figure \ref{ETC-fmnist-Modularity} shows that we achieve similar modularity as the baseline on the FMNIST-clustered dataset. The number of modules gradually converges to the same value as the baseline in Figure \ref{ETC-fmnist-Modules}. The result shows that our approach can adapt to the differences in the data distribution among clients. 

To further demonstrate the specialization of our approach. We test the approval pureness to quantify the ``specialized" of our approach on the FMNIST-clustered and Poets datasets, and the result is shown in Table \ref{pureness}. Since there are three clusters in the FMNIST-clustered dataset and two clusters of Goethe and Shakespeare in Poets, their basic pureness is 0.33 and 0.5, respectively. Our approach and the baseline's approach achieve 100\% pureness in the FMNIST-clustered data set, and all model approvals are from the same cluster. Our approach realizes the pureness similar to the baseline on the Poets dataset. The test result of the approval pureness shows the applicability of our approach in terms of implicit specialization for different datasets.

\begin{table}[ht]
    \centering
    \caption{The approval pureness after 100 rounds of training compared between our approach and baseline}
    \label{pureness}
    \begin{tabular}{cccccc}
        \toprule
        Dataset & \makecell[c]{Base\\Pureness} & \makecell[c]{Pureness\\of baseline} & \makecell[c]{Pureness\\of Ours}\\
        \midrule
        FMNIST-clustered & 0.33 & 1 & 1 \\
        Poets & 0.5 & 0.98 & 0.96 \\
        \bottomrule
    \end{tabular}
\end{table}

The above experiments show that our approach can achieve the implicit specialization of the ESDAGFL. Although in some metrics of the implicit specialization, the performance is not as good as the baseline, it is a negligible effect for the implicit specialization.

\subsubsection{Training Time Cost Evaluation}
The experiments in section \ref{train-acc-loss evaluation} and \ref{implicit-specialization evaluation} show that our approach realizes the balance between the training performance and specialization on FMNIST-clustered and Poets dataset. In this part, we will evaluate our approach's training time cost compared with the baseline. 

\begin{figure}[ht]
    \centering  
    \subfigure[FMNIST-clustered Dataset]{
        \label{ETC-fmnist-Duration}
        \includegraphics[width=80mm]{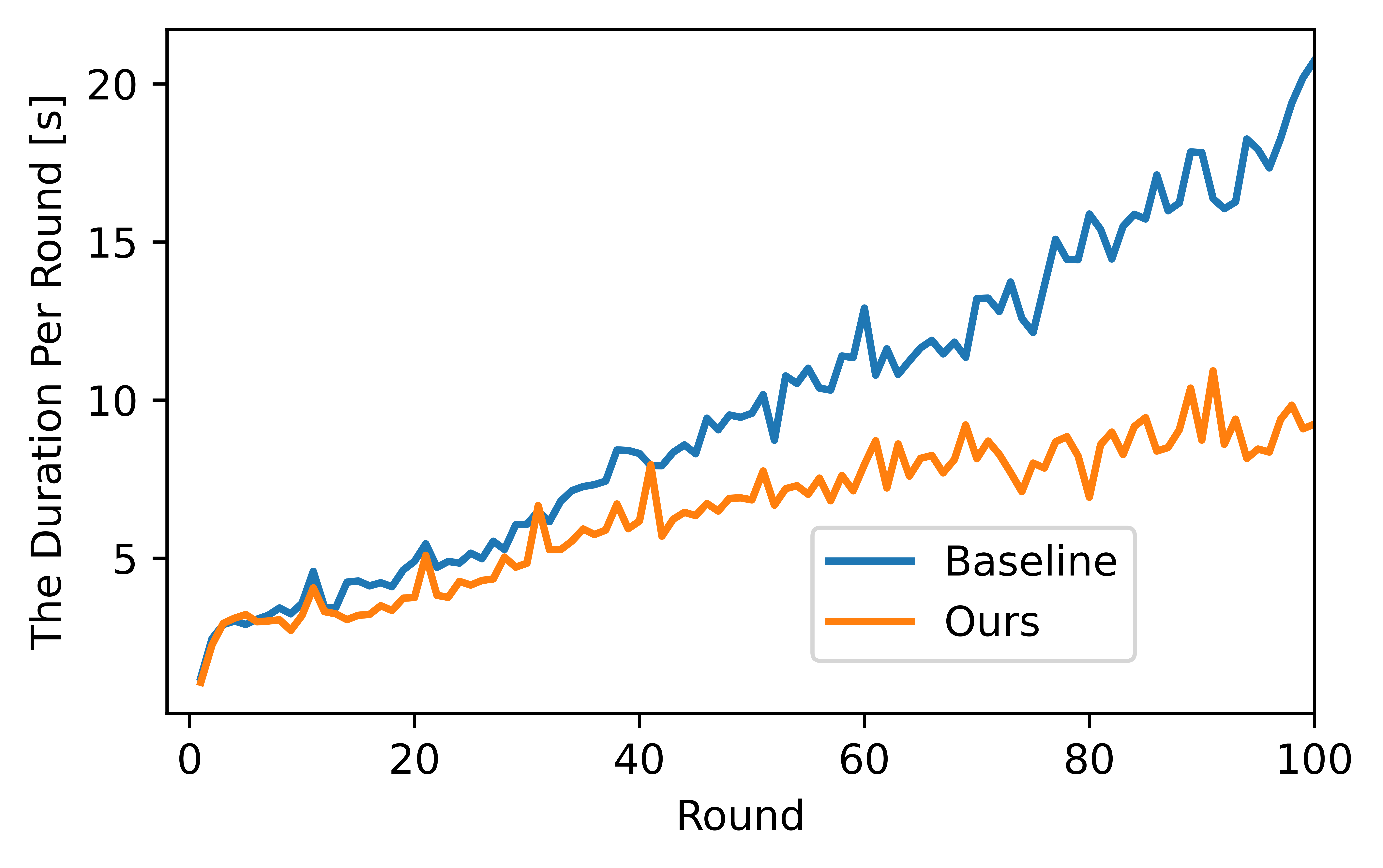}}
    \subfigure[Poets Dataset]{
        \label{ETC-Poet-Duration}
        \includegraphics[width=80mm]{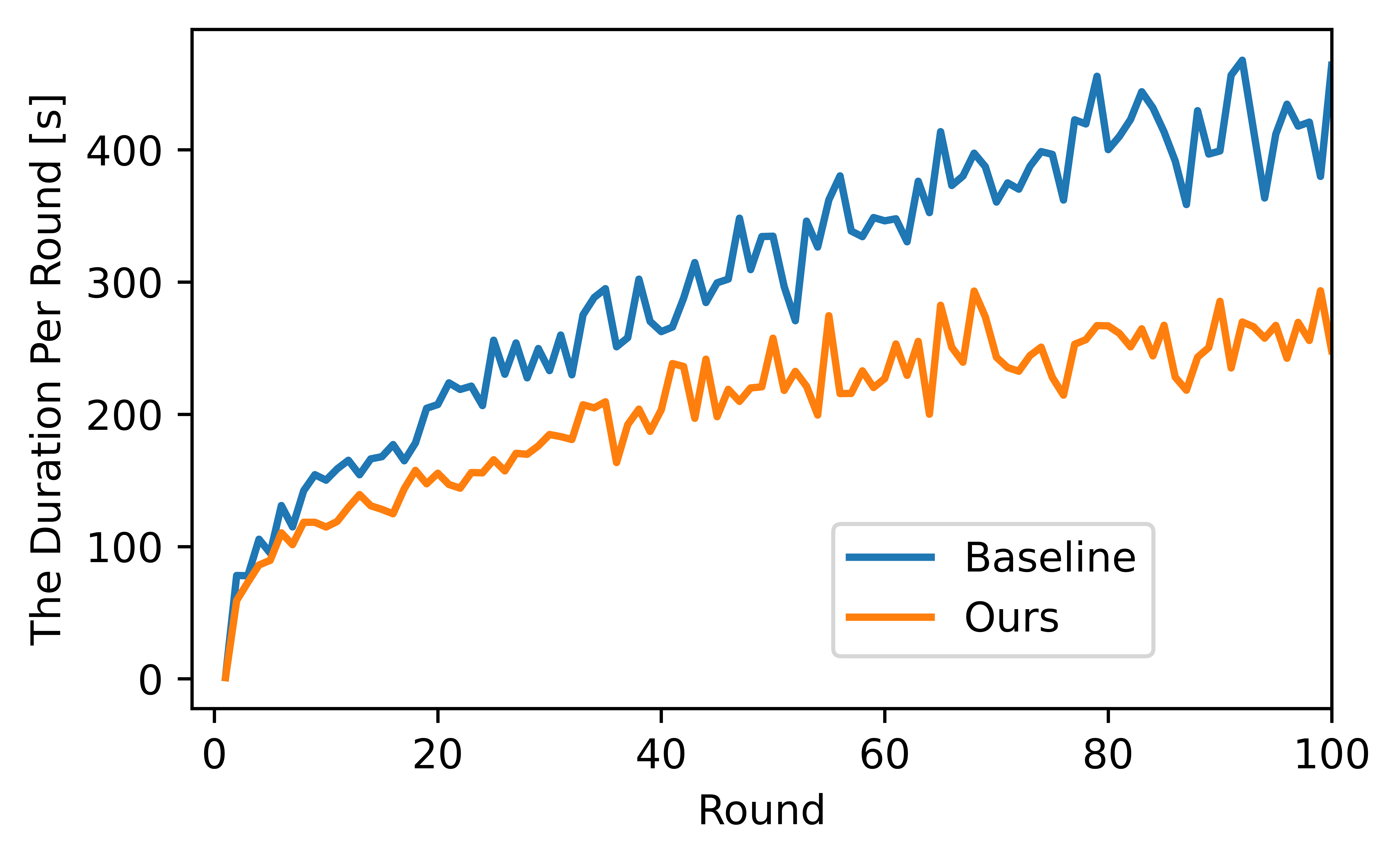}
        }
    \caption{Time cost in every round evaluated on FMNIST-clustered and Poets dataset.}
    \label{ETC-fmnist-Poets-Duration}
\end{figure}

Figure \ref{ETC-fmnist-Poets-Duration} shows the time cost per round of training during the simulation training process. The training time per round in baseline grows linearly. However, our approach's training time per round grows subconsciously and slowly. The time growth in our approach is mainly due to the time increases of the accuracy-biased random walk because of the increases in DAG-DLT. The total training time are reduced by 34\% and 33\% for the handwriting recognition task and next character prediction task, respectively.

As the energy consumption is directly proportional to the training time, we can conclude that the energy consumption of our approach is reduced by 34\% and 33\% for the handwriting recognition task and next character prediction task, respectively.
\section{Conclusion}
\label{sec: Conclusion}
Specializing DAG federated learning is a new paradigm of federated learning. In this work, we analyze the energy consumption of SDAGFL for the IoT scenario. Based on the formulated energy consumption analysis, we proposed an event-triggered communication-based SDAGFL approach, called ESDAGFL, to reduce the energy consumption of SDAGFL. The client participates in the ESDAGFL publishes the trained model only when it significantly changes from the averaged model. We evaluate our proposed approach on the FMNIST-clustered and Poets datasets. The experiments show that our method can realize reducing 33\% energy consumption compared with SDAGFL. It can reach an equal balance between the model's performance and specialization as SDAGFL. We will research the dynamic adjustment of trigger thresholds in ESDAGFL and communication overhead optimization in SDAGFL.



\bibliographystyle{elsarticle-num} 
\bibliography{references}





\end{document}